\documentclass[conference,letterpaper,10pt]{IEEEtran}
\IEEEoverridecommandlockouts

\usepackage[T1]{fontenc}
\usepackage[english]{babel}
\usepackage{textcomp}
\usepackage[table, pdftex, dvipsnames]{xcolor}
\usepackage{xspace}
\usepackage{graphicx}
\usepackage{amsfonts}
\usepackage{amsmath}
\usepackage{amssymb}
\usepackage{booktabs}
\usepackage[acronym]{glossaries}
\usepackage[capitalize,noabbrev]{cleveref} 
\usepackage{cite}
\usepackage{tikz}
\usepackage{multirow}
\usepackage{multicol}
\usepackage{colortbl}
\usepackage{siunitx}
\usepackage{algpseudocode}
\usepackage{algorithm}
\usepackage{url}
\usepackage[utf8]{inputenc}
\usepackage[most]{tcolorbox}
\usepackage{subcaption}
\glsdisablehyper
\AtBeginDocument{}
\tcbset{
    colback=gray!5,
    colframe=gray!40,
    left=6pt,
    right=6pt,
    top=6pt,
    bottom=6pt,
    boxsep=0pt,
}

\makeatletter
\DeclareRobustCommand\onedot{\futurelet\@let@token\@onedot}
\def\@onedot{\ifx\@let@token.\else.\null\fi\xspace}
\def\eg{\emph{e.g}\onedot} 
\def\ie{\emph{i.e}\onedot}

\makeatother

\begin{document}

\title{
\begin{tikzpicture}[overlay, remember picture]
    \node at ([yshift=-1.3cm]current page.north) {
        \normalsize\textcolor{gray}{This paper has been accepted for publication at the}
    };
    \node at ([yshift=-1.8cm]current page.north) {
        \normalsize\textcolor{gray}{17th ACM/IEEE International Conference on Cyber-Physical Systems (ICCPS), Saint Malo, France, 2026}
    };
\end{tikzpicture}

TinyML Enhances CubeSat Mission Capabilities
}

\author{
\IEEEauthorblockN{
Luigi Capogrosso\IEEEauthorrefmark{1},
Michele Magno\IEEEauthorrefmark{2}\IEEEauthorrefmark{1}}
\IEEEauthorblockA{
\IEEEauthorrefmark{1}Interdisciplinary Transformation University of Austria,
\IEEEauthorrefmark{2}ETH Zurich}
}
\maketitle

\begin{abstract}
Earth observation (EO) missions traditionally rely on transmitting raw or minimally processed imagery from satellites to ground stations for computationally intensive analysis.
This paradigm is infeasible for CubeSat systems due to stringent constraints on the onboard embedded processors, energy availability, and communication bandwidth.
To overcome these limitations, the paper presents a TinyML-based Convolutional Neural Networks (ConvNets) model optimization and deployment pipeline for onboard image classification, enabling accurate, energy-efficient, and hardware-aware inference under CubeSat-class constraints.
Our pipeline integrates structured iterative pruning, post-training INT8 quantization, and hardware-aware operator mapping to compress models and align them with the heterogeneous compute architecture of the STM32N6 microcontroller from STMicroelectronics.
This Microcontroller Unit (MCU) integrates a novel Arm Cortex-M55 core and a Neural-ART Neural Processing Unit (NPU), providing a realistic proxy for CubeSat onboard computers.
The paper evaluates the proposed approach on three EO benchmark datasets (\ie{}, EuroSAT, RS\_C11, MEDIC) and four models (\ie{}, SqueezeNet, MobileNetV3, EfficientNet, MCUNetV1).
We demonstrate an average reduction in RAM usage of 89.55\% and Flash memory of 70.09\% for the optimized models, significantly decreasing downlink bandwidth requirements while maintaining task-acceptable accuracy (with a drop ranging from 0.4 to 8.6 percentage points compared to the Float32 baseline). 
The energy consumption per inference ranges from 0.68 mJ to 6.45 mJ, with latency spanning from 3.22 ms to 30.38 ms.
These results fully satisfy the stringent energy budgets and real-time constraints required for efficient onboard EO processing.
\end{abstract}

\section{Introduction} \label{sec:intro}

Over the last decade, New Space \cite{Kodheli2021} has democratized access to orbit, making CubeSats, \ie{}, miniaturized satellites built from standardized $10 \times 10 \times 10$ cm units \cite{Liddle2020}, the core of many Earth Observation (EO) missions.
These missions span a variety of applications, including precision agriculture, maritime surveillance, and rapid response to disasters \cite{Crisp2020}.
However, CubeSats operate under tight constraints in size, weight, power consumption, and communication bandwidth that fundamentally stress the traditional EO operations model: acquire all sensor data, downlink them to ground stations, and process them on the ground or in the cloud \cite{Eosdis2025}.

\emph{\textbf{Motivations for this paper.}}
This conventional EO workflow was designed for large satellites with high‑capacity downlinks and continuous ground connectivity.
In this model, raw Level‑0 data is transmitted to ground stations and processed into higher‑level products in centralized facilities \cite{Eosdis2025}.
Although effective for flagship missions, this approach becomes impractical for CubeSats.
Typically, a low-Earth-orbit pass lasts only a few minutes, and even advanced S-band or X-band radios cannot compensate for the limited contact time and the energy cost per bit \cite{Babuscia2020}.
As a result, a large amount of collected data cannot be transmitted, or doing so incurs prohibitive delays and costs.

This bottleneck motivates a paradigm shift: instead of moving all data to the ground, we must bring computation to the data.
Onboard intelligence enables satellites to autonomously filter, classify, and prioritize information, reducing downlink volume and latency for time-critical applications.
In particular, recent missions such as $\Phi$‑Sat‑1 have demonstrated the feasibility of this approach, removing cloudy images in orbit and saving up to 30\% of downlink data \cite{Giuffrida2022}.

However, deploying Deep Neural Networks (DNNs) on CubeSat hardware is challenging because state‑of‑the‑art models require orders of magnitude more compute and memory than are available on typical CubeSat onboard computers \cite{Bayer2024}.
Specifically, commercial off-the-shelf onboard computers for 1U–3U CubeSats often use systems with only 16–64 MB of volatile memory and hundreds of MB of non-volatile storage, and must fit within power budgets of 2–8 W on average \cite{Bomani2021}.
These stringent resource envelopes render the direct deployment of uncompressed floating-point models, which are usually computationally and energy-intensive, infeasible under CubeSat-class resource constraints \cite{Diana2024}.

\emph{\textbf{Scientific contribution.}}
Achieving intelligence under the severe power, memory and compute constraints of CubeSat-class platforms requires a hardware-aware pipeline that jointly optimizes algorithmic choices with respect to the architecture of the target system \cite{Capogrosso2024}.
This work presents a TinyML-based Convolutional Neural Network (ConvNet) model optimization and deployment pipeline for onboard image classification, enabling accurate, energy-efficient, and fast inference on CubeSat-class hardware by integrating:
\begin{itemize}
    \item Structured iterative pruning to reduce model size and computational load.
    \item INT8 post-training quantization to compress weights and activations and leverage integer acceleration.
    \item Hardware-aware operator mapping to maximize utilization of heterogeneous computing resources.
\end{itemize}

Deployment experiments were conducted using a novel STM32N6 \cite{STM32N6}, specifically using the STM32N6570-DK, a modern Microcontroller Unit (MCU) platform that features an Arm Cortex‑M55 core at 800 MHz, 4.2 MB of SRAM to facilitate the handling of data-intensive Artificial Intelligence (AI) workloads, and an integrated Neural‑ART Neural Processing Unit (NPU) capable of delivering up to $\sim$600 GOPS for INT8 inference, enhancing processing capabilities crucial for real-time inference tasks.
Although this system is not physically implemented onboard a satellite, it serves as a representative scenario for designing a deployable inference pipeline within the strict limitations of CubeSat-class systems.
The use of STMicroelectronics devices in this context constitutes a practical and widely accepted approximation in the literature \cite{Dalbins2022, Abbas2024, Ghatul2024, Eshaq2025}.

To assess the effectiveness of our hardware-aware deployment pipeline, we conducted experiments on three EO-relevant datasets: EuroSAT \cite{Helber2019}, RS\_C11 \cite{Zhao2016}, and MEDIC \cite{Alam2022}, covering the tasks of land cover classification, risk assessment, and scene recognition.
Four representative models were evaluated, \ie{}, SqueezeNet \cite{Iandola2017}, MobileNetV3 \cite{Howard2019}, EfficientNet \cite{Tan2019}, and MCUNetV1 \cite{Lin2020}, focusing on three key metrics: classification accuracy, memory footprint, and energy consumption \cite{Capogrosso2026}.
We successfully demonstrate that this methodology compresses models from tens of megabytes to sub-megabyte footprints and reduces energy consumption to the millijoule scale.

In summary, the main contributions of this article are as follows.
\begin{itemize}
    \item First, we propose a systematic pipeline that combines pruning, quantization, and hardware‑aware operator mapping for embedded deep learning for the New Space.
    \item Second, we provide an experimental evaluation on a modern MCU + Neural-ART NPU platform, quantifying the accuracy drop (0.4–8.6 pp), latency (3.2–30.4 ms), energy consumption (0.68–6.45 mJ/inference), and memory trade-offs (average reductions of 89.55\% RAM and 70.09\% Flash) under CubeSat-class constraints.
    \item Finally, we demonstrate the generalizability of our pipeline across multiple EO-relevant datasets and model architectures for EO classification tasks.
\end{itemize}
Given the consistent resource savings and performance trade-offs observed, the underlying hardware-aware pipeline is extensible to other EO tasks such as semantic segmentation and object detection, which we identify as a direction for future work.
\section{Related Works} \label{sec:related}

\subsection{Onboard Deep Learning for Earth Observation}
In \cite{Giuffrida2020}, the authors introduce CloudScout, a ConvNet-based approach for onboard cloud detection in hyperspectral imagery.
The system eliminates cloud-contaminated data before downlinking, reducing bandwidth and energy usage.
The model is optimized for deployment on the Intel Myriad 2 VPU, achieving 92\% accuracy on the Sentinel-2 hyperspectral dataset, with an inference time of 325 ms, a power consumption of 1.8 W, and a footprint of 2.1 MB.
The mission $\Phi{}$‑Sat‑1, launched on 3 September 2020 to deploy the CloudScout model, marked a significant milestone as the first European satellite to operationalize AI for the triage of EO data \cite{Giuffrida2022}.

The OPS-SAT CubeSat, developed by the European Space Agency, explored the integration of model-based reasoning with machine learning to accelerate the data-to-decision cycle in orbit, proposing a framework that highlights the importance of model compression for inference efficiency \cite{Fratini2021}.

Another important mission is the KITSUNE 6U CubeSat from the Kyushu Institute of Technology, which includes a ConvNet for wildfire detection running on a Raspberry Pi on board the satellite \cite{Azami2021}.

In \cite{Vieilleville2020}, the authors used a knowledge distillation approach to compress DNNs for onboard EO image segmentation, focusing on ship detection.
In \cite{Murphy2023}, the authors analyze machine learning techniques for anomaly detection in satellite telemetry data.
The study categorizes anomalies and emphasizes the need for robust detection strategies to ensure satellite reliability and operational performance.
Finally, in \cite{Abbas2024}, the authors demonstrate how a trained ConvNet model was integrated into the AlAinSat-1 onboard STM32 MCU.

\subsection{TinyML on Microcontroller Units}
During the past five years, the TinyML community has researched optimized kernels \cite{Burrello2021}, frameworks \cite{Capogrosso2024}, and benchmarks \cite{Mattson2020a, Reddi2020, Capogrosso2026} to accelerate machine learning operations on MCUs.
For example, CMSIS‑NN provides efficient INT8 kernels for Arm Cortex‑M, providing $\sim$5$\times$ speed/energy gains over naive baselines \cite{Lai2018}.
TensorFlow Lite Micro provides a compact interpreter for MCU deployments with various vendors, with minimal runtime overhead \cite{Warden2019}.
Additionally, the MLPerf Tiny benchmark suite \cite{Mattson2020b} standardizes the assessment of accuracy, latency, and energy consumption across visual wake-word, image classification, and anomaly-detection workloads.

Advances in TinyML have also introduced automated neural architecture search and hardware-aware compression frameworks such as MCUNet \cite{Lin2020} and Once-for-All \cite{Cai2020}.
Other lightweight ConvNet families, such as SqueezeNet \cite{Iandola2017} and MobileNetV3 \cite{Howard2019}, have demonstrated strong efficiency on resource-constrained devices.

Finally, MCU vendors have also integrated neural accelerators to enhance on-device vision/audio processing.
Specifically, the STM32N6 from STMicroelectronics \cite{STM32N6} is an MCU that features a dedicated Neural-ART NPU designed for quantized ConvNets and is supported by the ST Edge AI Developer Cloud toolchain \cite{STEdgeAI}.
This toolchain partitions ONNX graphs \cite{ONNX}, offloads supported INT8 operations to the Neural-ART NPU, and reuses buffers and fused operations to reduce memory and latency.

\subsection{Gaps in the Literature}
Despite progress in enabling onboard intelligence for EO, several gaps remain unaddressed.

\emph{\textbf{Representative hardware constraints.}}
Although prior work has demonstrated onboard inference using embedded platforms, many rely on relatively powerful Systems on Chips (SoCs) or single-board computers (\eg{}, Intel Myriad, Raspberry Pi), which offer resources well beyond those available in typical CubeSat-class hardware.
This creates a gap in solutions explicitly designed for ultra-constrained environments where memory, compute, and energy budgets are extremely limited.

\emph{\textbf{End-to-end hardware-aware pipeline.}}
Although some model compression techniques, \eg{}, pruning, and quantization, have been investigated, there is a lack of systematic pipelines that jointly optimize models and deployment strategies.

\emph{\textbf{Comprehensive benchmarking.}}
Current studies rarely assess the trade-offs between energy, memory, and communication in realistic EO workloads, focusing mainly on accuracy or latency.
Standardized evaluations over multiple datasets remain limited.

\emph{\textbf{Addressing the gaps.}}
In this article, we directly address these gaps by first validating the solution under authentic ultra-constrained hardware constraints.
Second, we present a systematic end-to-end optimization and deployment pipeline.
Finally, we provide a comprehensive benchmark across multiple datasets and models, evaluating the complete trade-offs between RAM, Flash, energy consumption, and downlink bandwidth reduction. 
\section{Methodology} \label{sec:method}

The main contribution of this article is a lightweight yet accurate deep learning pipeline that operates within the strict computational, memory, and energy constraints of CubeSat-class hardware.
The following subsections detail the system design and the proposed methodology that systematically compresses and optimizes neural networks through three key stages: structured iterative pruning, post-training INT8 static quantization, and hardware-aware operator mapping.

\begin{figure}[t!]
    \centering
    \includegraphics[width=\linewidth]{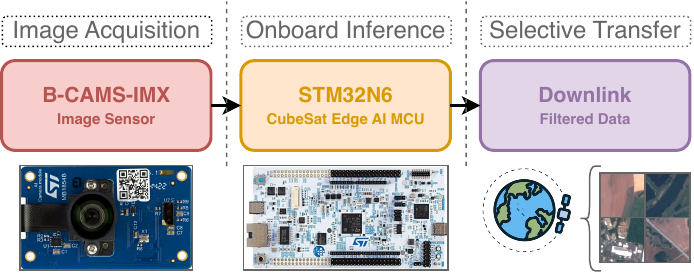}
    \caption{Overview of the STM32N6 prototyping environment for end-to-end Earth Observation workflows on ultra-low-power hardware devices.}
    \label{fig:overview}
\end{figure}
First, structured pruning removes redundant filters and neurons while preserving task performance, substantially reducing model complexity.
Next, static quantization further compresses the network by converting floating-point parameters and activations into 8-bit integer representations, minimizing both memory usage and computation cost.
Finally, a hardware-aware mapping process partitions the resulting model across the heterogeneous compute architecture of the STM32N6 platform, ensuring that operations are efficiently distributed between the Central Processing Unit (CPU) and the Neural-ART NPU.
Together, these steps form an integrated optimization pipeline (shown in \cref{fig:overview}) that enables the deployment of TinyML models for EO under realistic CubeSat constraints.

\subsection{System Design}

\emph{\textbf{STM32N6.}}
The STM32N6 is STMicroelectronics' flagship high-performance MCU, engineered specifically for TinyML applications.
It features an 800 MHz Arm Cortex-M55 core with Helium vector extensions, which enhances digital signal processing capabilities, crucial for real-time inference tasks.
A unique component is the integrated Neural-ART accelerator, ST’s proprietary NPU, which operates at up to 1 GHz, allowing efficient on-device execution of complex DNNs, as demonstrated by our EO tasks.
The MCU has 4.2 MB of embedded RAM, enabling efficient handling of data-intensive AI workloads.
The STM32N6 is supported by the ST Edge AI Suite, which includes tools such as STM32Cube.AI \cite{STM32CubeAI} and the ST Edge AI Developer Cloud \cite{STEdgeAI}, for the efficient deployment of learning models on ultra-low-power embedded systems.

\emph{\textbf{MB1854 B-CAMS-IMX.}}
This imaging subsystem features a high-resolution camera module designed for seamless integration with the STM32 ecosystem.
The processing of the captured frames is supported by the MCU’s integrated H.264 encoder.

\emph{\textbf{Communication and downlink constraints.}}
Standard telemetry is often transmitted on UHF or VHF frequencies, supporting low data rates of 1 to 9.6 kbps.
Although the S-band link offers a much higher data rate, up to 256 kbps, it can only be used intermittently when the satellite is over a ground station.
These communication opportunities are infrequent (only a few times a day) and are very brief, usually lasting around ten minutes \cite{Babuscia2020}.
Consequently, the total volume of data that can be transmitted daily is severely capped, often limited to just tens of megabytes.
This bottleneck makes onboard intelligent data handling essential.

\subsection{Design Goals}
We restricted our analysis to ConvNet or NAS-generated ConvNet models for two reasons.
First, the ST Edge AI toolchain currently provides optimized operator mapping primarily for convolutional architectures, limiting its compatibility with Transformer-based designs \cite{STEdgeAI}.
Second, focusing on ConvNets maximizes operator coverage in the Neural-ART NPU and simplifies integration into the deployment toolchain \cite{NPULimitations}.

Our selected models achieve high accuracy (see Section \ref{sec:experiments} for more details).
However, profiling on the STM32N6 with the ST Edge AI Suite reveals that the uncompressed Float32 network exceeds the on-chip RAM budget and violates the per-frame latency target required for real-time operation in most cases.
In short, the baselines are accurate but not deployable under CubeSat-class hardware constraints.

As a result, the optimization goals that guide our implementation are the following.
\begin{enumerate}
\item Limit the accuracy drop with respect to the Float32 baseline to a small and task-acceptable margin.
\item Fit the model parameters, intermediate activations, and working buffers within the 4.2 MB of embedded RAM.
\item Meet the per-frame inference deadline defined by the payload control loop on STM32N6.
\item Minimize energy consumption to extend mission life and flexibility in duty-cycling.
\item Reduce communication overhead by maximizing onboard inference.
\end{enumerate}

\subsection{Design Implications}

\emph{\textbf{Structured iterative pruning.}}
This step alternates between removing entire computational units, \ie{}, filters in convolutional layers or neurons in fully-connected layers, and fine-tuning the network based on the magnitude of their associated weights \cite{Capogrosso2024}, which we quantify using the per-layer $L_2$ Norm.

The procedure begins with a pre-trained model and iteratively applies the following pruning-fine-tuning cycle.
\begin{enumerate}
    \item \emph{Importance Ranking:} For each convolutional and linear layer, compute the $L_2$-norm for all of its output structures, \ie{}, filters or neurons.
    \item \emph{Pruning:} We rank the structures within each layer based on their importance scores and prune those with the lowest scores by applying a binary mask to the weight tensor.
    \item \emph{Fine-tuning:} Following the pruning step, fine-tune the pruned model for 10 epochs with a lower learning rate than the training one to allow the remaining weights to adjust and recover task performance.
\end{enumerate}

To mitigate the risk of severely damaging the network's learned representations, this cycle is not performed in a single aggressive step, but in a sequence of 3 stages with decreasing pruning rates.
Specifically, we prune the 10\% of the filters in the first iteration, then the 5\% in each of the two subsequent iterations.
This specific sequence was empirically determined to provide the optimal balance between significant initial compression and the model's ability to recover task performance during intermediate fine-tuning steps.
After the final iteration, the pruning masks are made permanent to yield the final pruned model.

\emph{\textbf{INT8 static quantization.}}
Following the pruning stage, we applied post-training static quantization to further compress the model and improve the inference efficiency \cite{Jacob2018}.
This method converts the 32-bit floating-point weights and activations into 8-bit signed integers (INT8).

Unlike dynamic quantization, static quantization requires a calibration step, in which the model is fed a small, representative dataset to determine the optimal quantization parameters for activations \cite{Hubara2021}.
The core of the quantization process is an affine mapping from a real value $r$ to its integer representation $q$.
The relationship is defined by a scale factor $S$ and a zero-point $Z$.
Specifically, $r = S \cdot (q - Z)$, where $S$ is a positive real number and $Z$ is an integer.
For INT8 quantization, $q$ is limited to the range $[-128, 127]$.
The parameters $S$ and $Z$ are determined during the calibration phase by observing the range $\left[\min(R), \max(R)\right]$, where $R$ is the set of real-valued tensor elements.

\emph{\textbf{Hardware-aware operator mapping.}}
Finally, we used the ST Edge AI Developer Cloud framework to perform hardware-aware operator mapping, \ie{}, partition the supported DNN operators to the Neural-ART NPU.
The mapping is performed at the operator level to maximize hardware acceleration and minimize latency.

Specifically, the tool inspects the quantized ONNX graph and partitions it into subgraphs based on hardware capabilities. 
Operators such as \texttt{Conv2D}, \texttt{DepthwiseConv2D}, and \texttt{Activations} (\eg{}, ReLU) are transferred to the Neural-ART NPU, which is optimized for INT8 arithmetic.
Other layers, such as custom or control-flow operations, remain on the CPU \cite{NPULimitations}.
To minimize latency and memory footprint, the optimizer applies:
\begin{itemize}
    \item \emph{Pipeline Scheduling:} Overlaps CPU and Neural-ART NPU execution where possible.
    \item \emph{Buffer Reuse:} Shares activation buffers across layers to reduce SRAM usage.
    \item \emph{Operator Fusion:} Combines compatible layers (\eg{}, \texttt{Conv} + \texttt{ReLU}) into single Neural-ART NPU kernels.
\end{itemize}
This hardware-aware step ensures that the model fully exploits the heterogeneous compute resources of the STM32N6 platform.
\section{Experiments} \label{sec:experiments}

\emph{\textbf{Datasets.}}
Our experimental evaluation is conducted on three different datasets: EuroSAT, MEDIC, and RS\_C11.
The selection of these datasets was guided by three main criteria.
Firstly, they are established as state-of-the-art benchmarks.
Secondly, they represent various EO tasks, including land cover classification (EuroSAT), remote sensing scene classification (RS\_C11), and risk assessment analysis (MEDIC).
Thirdly, they have significantly different image resolutions: $64 \times 64$ pixels for EuroSAT, $128 \times 128$ pixels for RS\_C11, and $224 \times 224$ pixels for MEDIC.

\emph{\textbf{Models.}}
We used the following models: SqueezeNet, MobileNetV3, EfficientNet, and MCUNetV1.
The MCUNetV1 model was excluded from the pruning phase because its architecture is already co-designed and optimized for resource-constrained devices, making further pruning unnecessary.

\emph{\textbf{Training details.}}
The models were implemented using PyTorch \cite{Paszke2019}.
For training dense models, we use the Adam optimizer \cite{Kingma2015} with a learning rate of $1 \times 10^{-3}$, a batch size of 32, and 50 epochs.
During the iterative pruning phase, the fine-tuning stage for each model utilized a reduced learning rate of $1 \times 10^{-4}$ and was conducted for 10 epochs.
All training and fine-tuning procedures were executed on a single NVIDIA H100 Tensor Core GPU.

\emph{\textbf{Evaluation setup.}}
We deployed models on the STM32N6570-DK discovery kit, using four key evaluation metrics: classification accuracy, inference latency, memory occupancy, and energy consumption.

\subsection{Accuracy and Memory Analysis}

\emph{\textbf{EuroSAT.}}
\begin{figure}[t!]
    \centering
    \includegraphics[width=\linewidth]{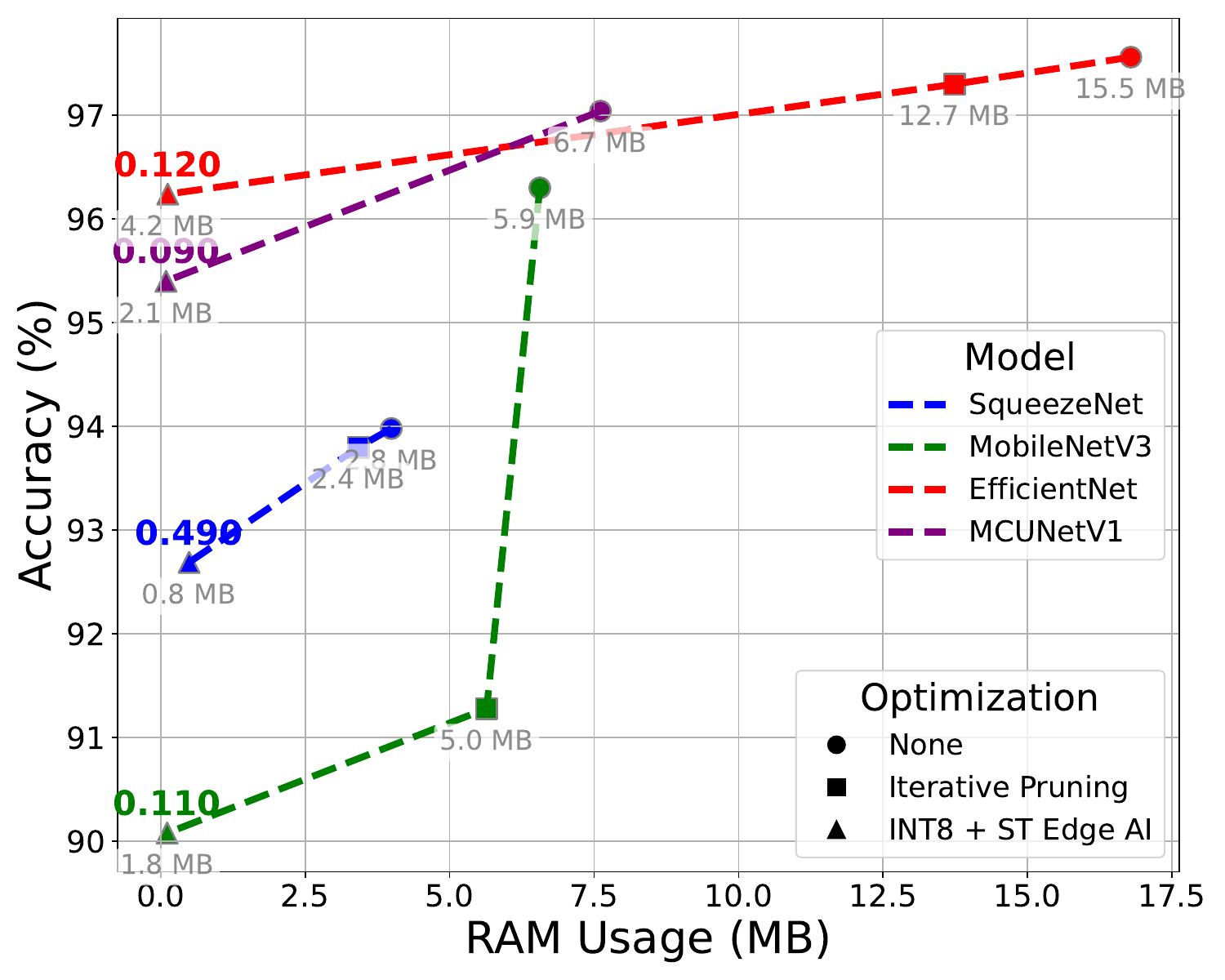}
    \caption{Quantitative results in terms of accuracy, RAM, and Flash (the grey numerical value reported below each data point) used on the EuroSAT dataset.}
    \label{fig:iccps-26-eurosat}
\end{figure}
The initial evaluation was conducted on the EuroSAT dataset ($64 \times 64$ images), and the results are presented in \cref{fig:iccps-26-eurosat}.
For baseline models, RAM usage was a significant barrier, with SqueezeNet, MobileNetV3, and EfficientNet requiring 3.99 MB, 6.56 MB, and 16.79 MB of RAM, respectively.

The optimized EfficientNet, which offered the highest baseline accuracy, achieved a final RAM footprint of 0.12 MB and a Flash footprint of 4.20 MB.
This represents a 99.30\% reduction in RAM and a 72.90\% reduction in Flash usage compared to the vanilla model.
These resource savings were achieved with only minor degradation in accuracy performance, with a reduction from 97.56\% to 96.24\%, \ie{}, a decrease of 1.32 percentage points.

The MobileNetV3 optimization yielded even more RAM savings.
The final model required only 0.11 MB of RAM, representing a 98.30\% reduction from its baseline of 6.56 MB.
This came at the cost of a relative decrease in accuracy of 6.22 percentage points.

SqueezeNet was compressed to 0.49 MB of RAM (a reduction of 87.70\%) and 0.82 MB of Flash (a reduction of 70.90\%), while its precision fell from 93.98\% to 92.69\%.

Finally, MCUNetV1, \ie{}, the most compact baseline, achieved a RAM footprint of 0.09 MB (a 98.80\% reduction from its 7.61 MB baseline) and a Flash footprint of 2.10 MB (a 68.60\% reduction from 6.69 MB).
This resource savings resulted in a decrease in accuracy from 97.04\% to 95.40\%, a decrease of 1.64 percentage points.

\emph{\textbf{RS\_C11 and MEDIC.}}
\begin{figure}[t!]
    \centering
    \begin{subfigure}[b]{0.49\linewidth}
        \centering
        \includegraphics[width=\linewidth]{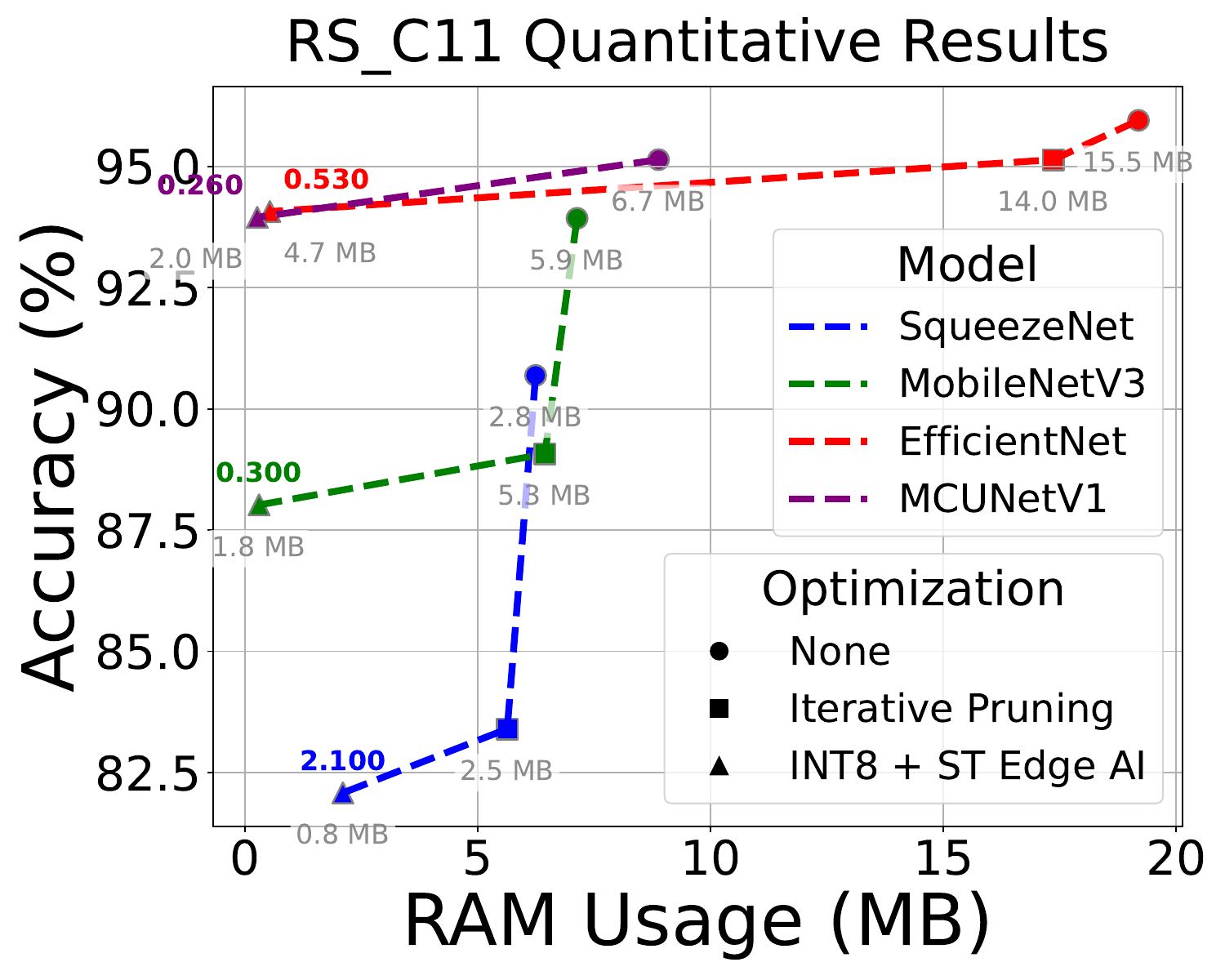}
    \end{subfigure}
    \hfill
    \begin{subfigure}[b]{0.49\linewidth}
        \centering
        \includegraphics[width=\linewidth]{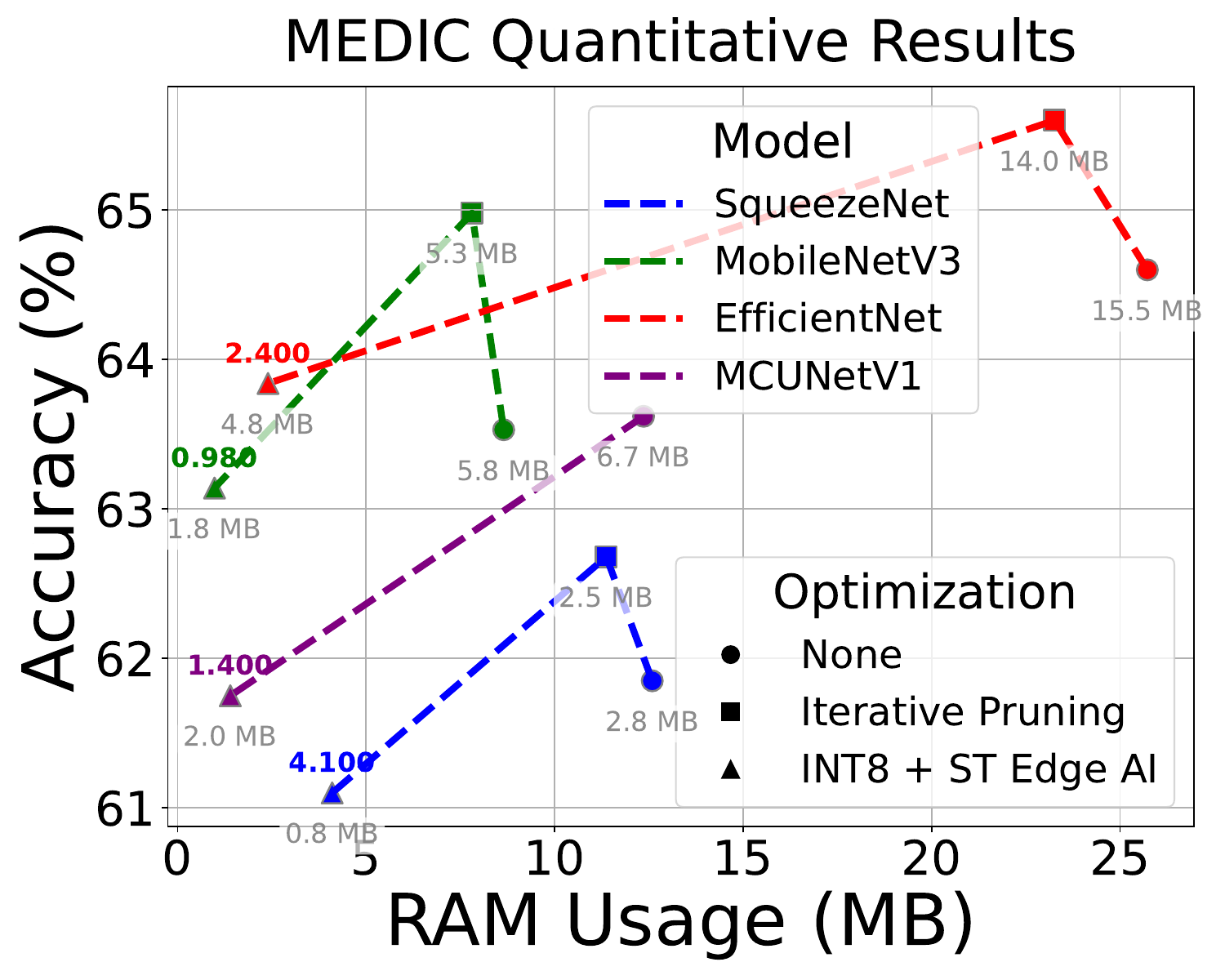}
    \end{subfigure}
    \caption{Quantitative results in terms of accuracy, RAM, and Flash (the grey numerical value reported below each data point) used on the RS\_C11 (left) and MEDIC (right) dataset.
    (*) Zoom in to clearly view the Flash values.}
    \label{fig:iccps-26-rsc_11-medic}
\end{figure}
We extended the experiments to two additional datasets: RS\_C11 ($128 \times 128$) and MEDIC ($224 \times 224$).
These datasets introduce higher complexity compared to EuroSAT, both in terms of image resolution and semantic variability.

\cref{fig:iccps-26-rsc_11-medic} (left) summarizes the results on RS\_C11.
The baseline models exhibited RAM footprints of 6.24 MB (SqueezeNet), 7.13 MB (MobileNetV3), and 19.19 MB (EfficientNet), values that clearly do not allow the deployment of these models on CubeSat-class hardware.
After applying iterative pruning, INT8 quantization and hardware-aware optimization, RAM usage decreased to 2.10 MB, 0.30 MB, and 0.53 MB, respectively, corresponding to reductions of 64.40\%, 95.80\%, and 97.20\%.
The MCUNetV1's RAM was further reduced to 0.26 MB.
Flash memory requirements were similarly compressed, with an average reduction of 70.10\% across all four models.
The resultant accuracy trade-off remains within acceptable bounds for many automated EO tasks, specifically: SqueezeNet decreased from 90.69\% to 82.08\%, MobileNetV3 from 93.93\% to 88.02\%, EfficientNet from 95.95\% to 94.07\%, and MCUNetV1 from 95.14\% to 93.95\%.

MEDIC, with its resolution of $224 \times 224$, represented the most computationally demanding scenario, testing the effectiveness of the hardware-aware mapping in balancing resource constraints with task complexity.
As shown in \cref{fig:iccps-26-rsc_11-medic} (right), optimized models achieved RAM footprints of 4.10 MB (SqueezeNet), 1.10 MB (MobileNetV3), 2.40 MB (EfficientNet), and 1.40 MB (MCUNetV1).
This compares to baseline values that, for the three standard models, exceeded 6 MB (12.59 MB, 8.65 MB, and 25.72 MB, respectively).
Even MCUNetV1 had a significant reduction from 12.36 MB to 1.40 MB.
Reductions in flash usage were equally significant.
It is important to note an interesting phenomenon that occurs with this specific dataset: iterative pruning consistently improves the performance of all baseline models.
Specifically, the accuracy of SqueezeNet increases from 61.85\% to 62.68\%, MobileNetV3 increases from 63.53\% to 64.98\%, and EfficientNet exhibits a jump from 64.60\% to 65.60\% after pruning.
This improvement occurs because the pruning process acts as a form of regularization, effectively removing redundant parameters and noise that might lead to overfitting, allowing the models to generalize better on the data.
Unlike the EuroSAT and RS\_C11 datasets, where the models likely already achieved a near-optimal fit, this dataset presents a specific distribution where sparsification helps the network focus on more robust features.

\textcolor{blue}{\emph{Discussion.}}
The optimized models maintain a task-acceptable accuracy margin compared to their Float32 baselines, with most accuracy drops remaining below 10 percentage points, thus satisfying the design goal \#1.
At the same time, memory footprint reductions in RAM (and in Flash) ensure that all model parameters, activations, and buffers fit within the 4.2 MB embedded RAM of the STM32N6, fulfilling the design goal \#2.

\subsection{Inference and Power Consumption Analysis}

\begin{figure}[t!]
    \centering
    \includegraphics[width=\linewidth]{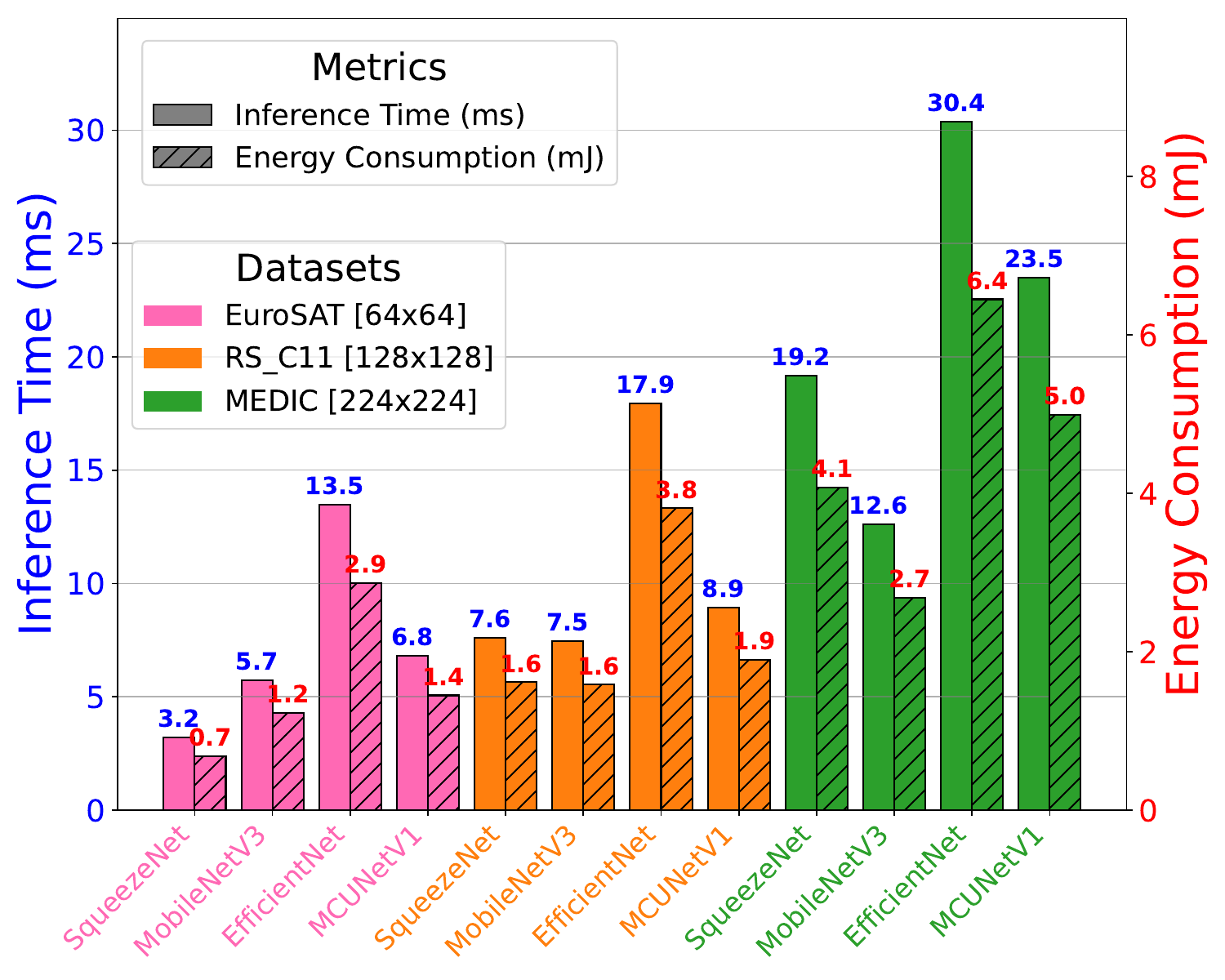}
    \caption{Inference latency and energy consumption for fully optimized models across all the datasets.
    Left: Latency in milliseconds; Right: Energy per inference in millijoules.}
    \label{fig:iccps-26-inference-energy}
\end{figure}
We measured inference time and energy consumption for fully quantized models, with results summarized in \cref{fig:iccps-26-inference-energy}.

As expected, energy consumption scales with the resolution of the input image.
On EuroSAT, the lightweight SqueezeNet was the most efficient, consuming only 0.68 mJ per inference.
MobileNetV3, MCUNetV1, and EfficientNet consumed 1.23 mJ, 1.45 mJ, and 2.86 mJ, respectively.
On RS\_C11, the energy cost increased for all models, requiring 1.61 mJ (SqueezeNet), 1.58 mJ (MobileNetV3), 1.90 mJ (MCUNetV1), and 3.81 mJ (EfficientNet).
Finally, on MEDIC, the dataset with the highest resolution, we observe the highest energy draw, with SqueezeNet consuming 4.07 mJ, MobileNetV3 consuming 2.68 mJ, MCUNetV1 consuming 4.99 mJ, and EfficientNet consuming 6.45 mJ.

\textcolor{blue}{\emph{Discussion.}}
These results confirm compliance with the design goals \#3 and \#4.
All optimized models operate within milliJoule energy budgets, with the most efficient configuration (SqueezeNet on EuroSAT) consuming only 0.68 mJ per inference.
Even in the most demanding scenario (EfficientNet on MEDIC), the energy cost remains within feasible limits for CubeSat-class hardware.
The inference latencies achieved range from 3.20 ms to 30.40 ms (\ie{}, $\sim$312.5 FPS to $\sim$32.9 FPS).
This range demonstrates that all configurations meet or exceed the real-time requirement of 5 FPS for payload control in low Earth orbit imaging \cite{Mellinkoff2017}.

\subsection{Communication and Downlink Analysis}
To quantify the impact of onboard inference on communication, we analyze the cost of transmitting all samples to a ground station versus selectively transmitting only those samples for which the onboard model exhibits low confidence.

Consider the EuroSAT dataset.
Each image, with a resolution of $64 \times 64$ pixels and three color channels, occupies approximately 12.3 KB of storage space.
As a result, transmitting all 5,400 images from the test set would require $\sim$66.4 MB of bandwidth.
This volume exceeds the typical daily transmission budget for CubeSat missions.
For higher-resolution datasets, such as MEDIC, the raw data volume would be even more prohibitive.

In contrast, let us consider a hybrid inference scenario in which the onboard model performs classification and transmits only those samples for which its confidence is below a predefined threshold (\ie{}, 95\%).
These uncertain samples are then reprocessed on the ground using a more powerful model (in our case, the non-optimized baseline).
For example, using MobileNetV3 onboard, only 14.23\% of the EuroSAT images (\ie{}, 768 out of 5,400) fall below the confidence threshold and are transmitted.
This results in a total transmission volume of approximately 9.45 MB, representing an 85.77\% reduction in communication load compared to full dataset transmission.

\textcolor{blue}{\emph{Discussion.}}
The advantages of this hybrid approach are twofold.
First, it achieves a 85.77\% reduction in data traffic compared to baseline, successfully meeting the design goal \#5 by fitting well within the strict daily communication budget.
Second, this method improves the total performance of the system.
Although MobileNetV3 alone achieves an accuracy of 90.08\%, the hybrid system, which uses a stronger ground-based model for uncertain cases, achieves a combined final accuracy of 95.20\%, demonstrating that onboard inference, when combined with confidence-based filtering, can simultaneously optimize communication efficiency and system-level accuracy.
\section{Concluding Remarks} \label{sec:conclusions}

This paper addressed the critical challenge of deploying advanced EO models on CubeSats, platforms severely limited by stringent constraints on power, memory, and communication bandwidth.

We presented a TinyML pipeline that integrates structured iterative pruning, post-training INT8 quantization, and hardware-aware operator mapping, evaluated on the STM32N6 board, a representative hardware proxy for CubeSat-class systems.

The experimental results demonstrate the effectiveness of the pipeline using multiple models (\ie{}, SqueezeNet, MobileNetV3, EfficientNet, MCUNetV1) on three distinct EO datasets (\ie{}, EuroSAT, RS\_C11, MEDIC).
We successfully compressed multi-megabyte floating-point models into deployable solutions while maintaining acceptable trade-offs in accuracy.
Furthermore, all optimized models operate with millijoule-scale energy consumption and meet real-time performance requirements.
Finally, we demonstrate that onboard intelligence can reduce downlink bandwidth requirements while simultaneously improving overall system accuracy.

\bibliographystyle{IEEEtran}
\bibliography{bibliography}

\end{document}